\pdfoutput=1

\documentclass{article}
\usepackage{ijcai25}

\usepackage{times}
\usepackage{soul}
\usepackage{url}
\usepackage[hidelinks]{hyperref}
\usepackage[utf8]{inputenc}
\usepackage[small]{caption}
\usepackage{graphicx}
\usepackage{amsmath}
\usepackage{amsthm}
\usepackage{booktabs}
\usepackage{algorithm}
\usepackage{algorithmic}
\usepackage[switch]{lineno}

\usepackage{dsfont}
\usepackage{amsmath}
\usepackage{newfloat}
\usepackage{listings}
\usepackage{colortbl}
\usepackage{multirow}
\usepackage{pifont}
\usepackage{booktabs}
\usepackage[utf8]{inputenc}
\usepackage{url}
\usepackage{booktabs}
\usepackage{amssymb}
\usepackage{bbding}
\usepackage{pifont}
\usepackage{wasysym}
\usepackage{utfsym}
\usepackage{fontawesome}
\usepackage{pifont}       
\usepackage{bbding}       
\usepackage{fontawesome}
\newcommand{\cmark}{\ding{51}}%
\newcommand{\xmark}{\ding{55}}%
\usepackage{float}
\usepackage{bm}
\usepackage{tikz}

\newcommand{\ie}{\textit{i}.\textit{e}.}
\newcommand{\eg}{\textit{e}.\textit{g}.}

\newcommand{\testv}{\textbf{\textsc{TeST-V}}}
\usepackage{newfloat}
\usepackage{listings}

\usepackage{tablefootnote}

\title{${\textsc{TeST-V}}$: TEst-time Support-set Tuning for Zero-shot Video Classification}

\author{
Rui Yan$^1$
\and
Jin Wang$^2$
\and
Hongyu Qu$^2$
\and
Xiaoyu Du$^2$
\and
Dong Zhang$^3$
\and\\
Jinhui Tang$^{2}$\And
Tieniu Tan$^1$\\
\affiliations
$^1$Nanjing University\\
$^2$Nanjing University of Science and Technology\\
$^3$Hong Kong University of Science and Technology\\
\emails
ruiyan@nju.edu.cn,
wangjin-andy@njust.edu.cn,
\{quhongyu, duxy\}@njust.edu.cn,
dongz@ust.hk,
jinhuitang@njust.edu.cn,
tnt@nju.edu.cn
}

\begin{document}

\maketitle

\begin{abstract}
Recently, adapting Vision Language Models (VLMs) to zero-shot visual classification by tuning class embedding with a few prompts (Test-time Prompt Tuning, TPT) or replacing class names with generated visual samples (support-set) has shown promising results. However, TPT cannot avoid the semantic gap between modalities while the support-set cannot be tuned. To this end, we draw on each other's strengths and propose a novel framework namely $\textbf{TE}$st-time $\textbf{S}$upport-set $\textbf{T}$uning for zero-shot $\textbf{V}$ideo Classification ($\textbf{TEST-V}$). It first dilates the support-set with multiple prompts (Multi-prompting Support-set Dilation, MSD) and then erodes the support-set via learnable weights to mine key cues dynamically (Temporal-aware Support-set Erosion, TSE). Specifically, $\textbf{i) MSD}$ expands the support samples for each class based on multiple prompts enquired from LLMs to enrich the diversity of the support-set. $\textbf{ii) TSE}$ tunes the support-set with factorized learnable weights according to the temporal prediction consistency in a self-supervised manner to dig pivotal supporting cues for each class. $\textbf{TEST-V}$ achieves state-of-the-art results across four benchmarks and has good interpretability for the support-set dilation and erosion.
\end{abstract}
\section{Introduction}
\label{intro}
\begin{figure*}[t]
  \centering
\includegraphics[width=0.95\linewidth]{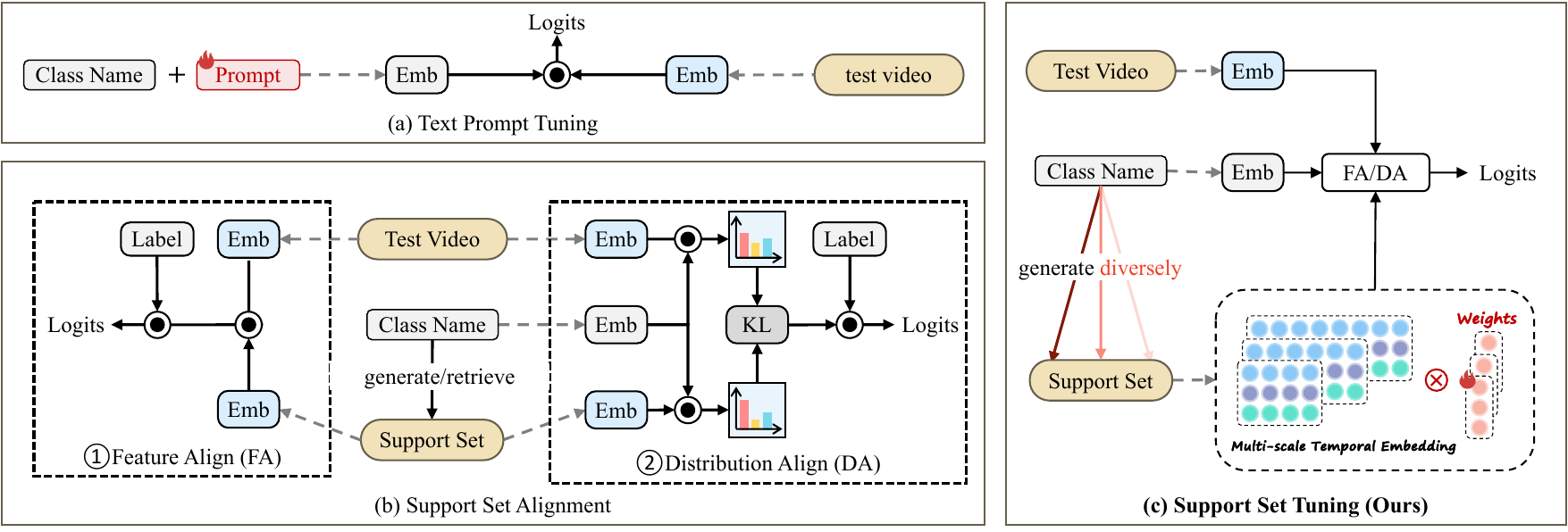}
  \caption{Innovation of the zero-shot activity recognition framework. a) Tuning the text input via the given test video in a self-supervised manner. b) Aligning the test video with the label based on the support set from feature similarity or predicted distribution similarity. c) This work combines the above thoughts to construct the support set diversely and tunes this set in a self-supervised manner to mine high-quality support samples.}~\label{fig::mov}
\end{figure*}

{
In the past few decades, the research on behavior recognition has made rapid progress and has been successfully applied in many fields, including intelligent robotics, security surveillance, and automatic driving. However, the number of behaviors existing methods can identify is still limited, which makes them difficult to meet the growing needs of practical applications. In recent years, the powerful ability of large-scale pre-trained Vision-Language models~(VLMs) in zero-shot generalization has promoted the rapid development of zero-shot/general behavior recognition. VLMs bring hope for breaking through the perceptual boundaries of existing behavior recognition technologies. 

Recently, it has become a trend to project the test video and class names into the joint feature space based on the pre-trained models, and then assign a label to the video according to their feature similarities. Although the amount of pre-training data is huge, the out-of-distribution issue cannot be thoroughly eliminated in the test environment. Hence, it is necessary to further tune the parameters or prompts for the pre-trained model from different modal perspectives.

To address this challenge, existing solutions can be roughly divided into the following two categories. i)~\textbf{Training-based}: Conventional methods fine-tune \textit{part of parameters} belonging/appended to the image-based pre-trained model~(\eg, CLIP~\cite{radford2021learning}) or \textit{extra learnable prompts} appended to (visual/text) inputs, based on a mass of video data. ii)~\textbf{Training-free}: To avoid high training costs, some recent Test-time Prompt Tuning~(TPT)-based methods~\cite{shu2022tpt,yan_2024_DTSTPT} append learnable prompts to visual/text input sequence and tune prompts with a self-supervised manner~(\eg, multi-view prediction consistency) based on the single test video during test time~(Figure~\ref{fig::mov}~(a)). Beyond that, some recent works~\cite{udandarao2022sus-x,zhang2021tipadaptertrainingfreeclipadapterbetter} build a video support-set according to unseen class names via retrieval or generation(\ie, converts class names to videos) and performs intra-modality (video-video) alignment (Figure~\ref{fig::mov}~(b)). TPT freezes the visual space and only tunes the text space, which cannot reduce the modality gap. On the contrary, the support-set-based methods reduce the gap via inter-modality alignment but the supporting samples are fixed.

}

{To this end, we proposed a novel framework namely \textbf{TE}st-time \textbf{S}upport-set \textbf{T}uning for \textbf{V}ideo classification (\testv) in Figure~\ref{fig::mov}~(c). It builds a semantically diverse support-set via off-the-shelf text-video generation models and tunes learnable weights to select effective support samples from the set via two core modules. \textbf{i) Multi-prompting Support-set Dilation~(MSD)}: generate multiple text prompts from the class names via LLMs and feed them into the text-video generation model~(\ie, LaVie~\cite{wang2023laviehighqualityvideogeneration}) for building a semantically diverse support set. \textbf{ii) Temporal-aware Support-set Erosion~(TSE)}: tunes a few learnable weights to dynamically reshape the contribution of each video frame in a self-supervised manner, according to the multi-scale semantic consistency hypothesis in the temporal domain. Extensive experimental results show that \textbf{TEST-V} improves the state-of-art pre-trained VLMs, namely CLIP~\cite{radford2021learning}, BIKE~\cite{wu2023bidirectional} and VIFi-CLIP~\cite{hanoonavificlip} by {$2.98\%$}, $2.15\%$, and $1.83\%$ absolute average accuracy respectively across four benchmarks. 
}

{Our main contributions can be summarized as follows:
\begin{itemize}
    \item We propose a novel framework, namely \textbf{TE}st-time \textbf{S}upport-set \textbf{T}uning which first dilates and then erodes the visual support-set to enhance the zero-shot generalization on activity recognition during test time.
    \item To ensure the diversity of the support set, we propose a Multi-prompting Support-set Dilation~(MSD) module to generate videos via text-to-video model as supporting samples for each class according to multiple descriptions.
    \item To mine critical supporting cues from the support-set, we proposed a Temporal-aware Support-set Erosion~(TSE) module to tune factorized learnable weights with different temporal-scale features in multiple steps supervised by the prediction consistency.
\end{itemize}
}
\section{Related Work}
\label{Related Work}
\subsection{Zero-shot Activity Recognition}
Zero-Shot Activity Recognition (ZSAR) aims to recognize actions that are not observed by the model during training, which is a non-trivial task yet useful for practical applications. Early work learn a joint embedding space by aligning video representation with corresponding textual semantic representation, \eg, manually defined attributes~\cite{liu2011recognizing,zellers2017zero}, word embeddings of category names~\cite{qin2017zero,shao2020temporal}, and elaborative action descriptions~\cite{chen2021elaborative,qian2022rethinking}. With the advent of deep neural networks, many works~\cite{chen2021elaborative,lin2019tsm,lin2022cross} make use of various video architectures (\eg, I3D~\cite{carreira2017quo} and TSM~\cite{lin2019tsm}) to obtain high-level visual representation from videos and further project such visual representation into semantic embedding space. Recently, large-scale Vision-Language models (\eg, CLIP~\cite{radford2021learning}) have shown great zero-shot transfer ability. Thus many ZSAR solutions attempt to take the advent of CLIP to align video features and textual features, showing impressive performance in ZSAR. ActionCLIP~\cite{wang2021actionclip}, A5~\cite{ju2022prompting}, and XCLIP~\cite{XCLIP}  adapt CLIP for videos by training additional components, \eg, temporal transformer blocks and text prompts. ViFi-CLIP~\cite{hanoonavificlip}, BIKE~\cite{wu2023bidirectional}, and MAXI~\cite{lin2023match} fully fine-tune the encoder of CLIP to learn video-specific inductive biases, without introducing any additional learnable parameters. Different from current approaches that rely heavily on many labeled video samples to fine-tune network parameters, our work pioneers the idea of building an efficient visual support set for zero-shot classification without training. 

\subsection{Adaptation of Visual-Language Models}
Recently, pre-trained Vision-Language Models (VLMs) (\eg, CLIP~\cite{radford2021learning}, ALIGN~\cite{jia2021scaling}) have shown impressive zero-shot transfer ability in recognizing novel categories. Thus many works attempt to adapt VLMs for downstream tasks through prompt tuning or training-free methods. CoOp~\cite{zhou2022learning} fine-tunes learnable text prompts on downstream training data instead of using hand-crafted templates~\cite{zhou2022learning}, effectively improving CLIP zero-shot performance. CoCoOp~\cite{zhou2022conditional} extends CoOp to learn input-conditioned prompts, leading to better generalizing to unseen classes and different domains. Despite impressive, these works need access to labeled samples from target distribution, and further train prompts over these samples. In addition, recent training-free methods such as TIP-Adapter~\cite{zhang2021tipadaptertrainingfreeclipadapterbetter} leverage a few labeled training samples as a key-value cache to make zero-shot predictions during inference, avoiding the conventional model parameter fine-tuning. Yet these methods still rely on samples from the target distribution. To discard costly labeled data and training, SuS-X~\cite{udandarao2022sus-x} utilizes category name and text-to-image generation model to construct visual support-set, thus enhancing zero-shot transfer abilities. Our test-time support-set tuning shares a similar spirit of pursuing a visual support-set curation strategy to adapt VLMs, but we strive to construct a diverse video support set guided by elaborated descriptions of each class name, and make full use of each sample in the support-set by dynamically adjusting the contribution of each frame.

\subsection{Test-time Tuning}
Test-time Tuning (TTT)~\cite{shocher2018zero,nitzan2022mystyle,xie2023sepico} aims to improve the generalization capabilities of pre-trained models by utilizing test samples to fine-tune the model, especially when faced with data distribution shifts. One of the key challenges in TTT is designing effective optimization objectives. Early works~\cite{sun2020test,liu2021ttt} enhance the training process by integrating a self-supervised multitask loss. Besides, TENT~\cite{wang2020tent} minimizes predictions' entropy on target data to improve generalization capability, yet requiring multiple test samples. To address this issue, MEMO~\cite{zhang2022memo} and TPT~\cite{shu2022tpt} utilize multiple augmented views generated from a single test sample through data augmentation for further test-time tuning. Recent methods extend TTT to the video domain by self-supervised dense tracking~\cite{azimi2022self} or aligning video features with different sampling rates~\cite{zeng2023exploring}. Inspired by this, we tune the support-set via learnable parameters with temporal prediction consistency loss for mining reliable supporting cues.

\section{Preliminary and Definition}
\label{sec:method}

\subsection{Problem Statement}
Zero-Shot Activity Recognition~(ZSAR) aims to recognize unseen actions during model testing. Specifically, given a test video $\bm{V}_\mathrm{test}$ and a set of class names $\bm{Y}=\{y_{0},y_{1},\cdots,y_{c}\}$, we employ the pre-trained visual~($\mathcal{F}_\mathrm{vis}$) and text~($\mathcal{F}_\mathrm{txt}$) encoders~\emph{e.g.}, CLIP~\cite{radford2021learning}, to encode them and then calculate the feature similarity for prediction as follows,
\begin{equation}
\begin{gathered}
\bm{f} = \mathcal{F}_\mathrm{vis}(\bm{V}_\mathrm{test}), \bm{W} = \mathcal{F}_\mathrm{txt}(\bm{Y}),\\
\bm{Z}_{\mathrm{ZS}} = \bm{f} \cdot \bm{W}^\mathrm{T}, \bm{f} \in \mathbb{R}^{d}, \bm{W} \in \mathbb{R}^{C \times d}.
\end{gathered}
\end{equation}
Here, $\bm{f}$ and $\bm{W}$ denote visual and text features respectively, and ``$\cdot$" calculates their cosine similarity.

Although the pre-trained representation is robust enough, it cannot avoid the inherent semantic gap between visual and text features~(\emph{a.k.a}, modality gap). To this end, some recent methods (\emph{e.g.}, TIP-Adapter~\cite{zhang2021tipadaptertrainingfreeclipadapterbetter}, SuS-X~\cite{udandarao2022sus-x}) attempt collecting a set of visual samples (support-set) as the replacement of class names and perform zero-shot matching in the visual space for reduce such gap. We first briefly review such support-set-based zero-shot matching methods as below.

\subsection{Support-Set-based Zero-shot Prediction}
\noindent\textbf{Support-Set Construction.}~Given a set of class name $\bm{Y}$, the text-to-image (\texttt{T2I}) generation model (\emph{e.g.}, Stable Diffusion~\cite{rombach2022high}) is employed to generate $K$ images for each category to construct the support set. To obtain more diverse generated samples, SuS-X~\cite{udandarao2022sus-x} explains each class name in detail through LLMs~(\eg, GPT-3~\cite{brown2020languagemodelsfewshotlearners}) as input of \texttt{T2I} model following CuPL~\cite{pratt2023does}. Thus, this support-set consists of $C \times K$ visual samples where $C$ and $K$ denote the number of classes and the number of generated samples for each class.

\noindent\textbf{TIP-Adapter Inference.}
Based on the above support-set, CLIP's visual encoder $\bm{E}_\mathrm{v}$ is applied to extract its visual features $\bm{F} \in \mathbb{R}^{CK \times d}$, and its labels are converted into the one-hot vector $ \bm{L} \in \mathbb{R}^{CK \times C}$. TIP-Adapter aims to calculate the distance between the test video feature $\bm{f}$ and support videos' feature $\bm{F}$, and then make predictions~($\bm{Z}_{\mathrm{TA}}$) with the help of label information $L$ as follows,
\begin{equation}
\bm{Z}_{\mathrm{TA}} =\exp (-\beta (1 - \bm{f}\bm{F}^\mathrm{T})) \bm{L}. 
\end{equation}
Here, $\beta$ adjusts the sharpness and affinities calculated from $\bm{f}\bm{F}^\mathrm{T}$ is used as attention weights over $\bm{L}$ to achieve logits.

\noindent\textbf{TIP-X Inference.} To avoid the uncalibrated intra-modal feature distances in TIP-Adapter, TIP-X~\cite{udandarao2022sus-x} uses text modality as a bridge between visual modalities. TIP-X calculates the distance between the test video feature $f$ and text features $\bm{W}$, and between support videos' feature $F$ and text feature $W$, respectively. TIP-X applies KL-divergence to construct the affinity matrix for final prediction~($\bm{Z}_{\mathrm{TX}}$) as follows,
\begin{equation}
\bm{Z}_{\mathrm{TX}}=\psi (-\mathtt{KL}(\bm{f} \bm{W}^\mathrm{T} || \bm{F}\bm{W}^\mathrm{T})) \bm{L}.\label{eq::TIP-X}
\end{equation}
Here, $\bm{f} \bm{W}^\mathrm{T}$ and $\bm{F}\bm{W}^\mathtt{T}$ are normalized via softmax, $\mathtt{KL}(\cdot)$ computes the similarity between two probabilities~($\mathtt{KL}(\bm{P} \parallel \bm{Q}) = \sum_i \bm{P}_i \log \frac{\bm{P}_i}{\bm{Q}_i}$) and $\psi$ is the scaling factor.

\begin{figure*}[t]
  \centering
  \includegraphics[width=0.95\linewidth]{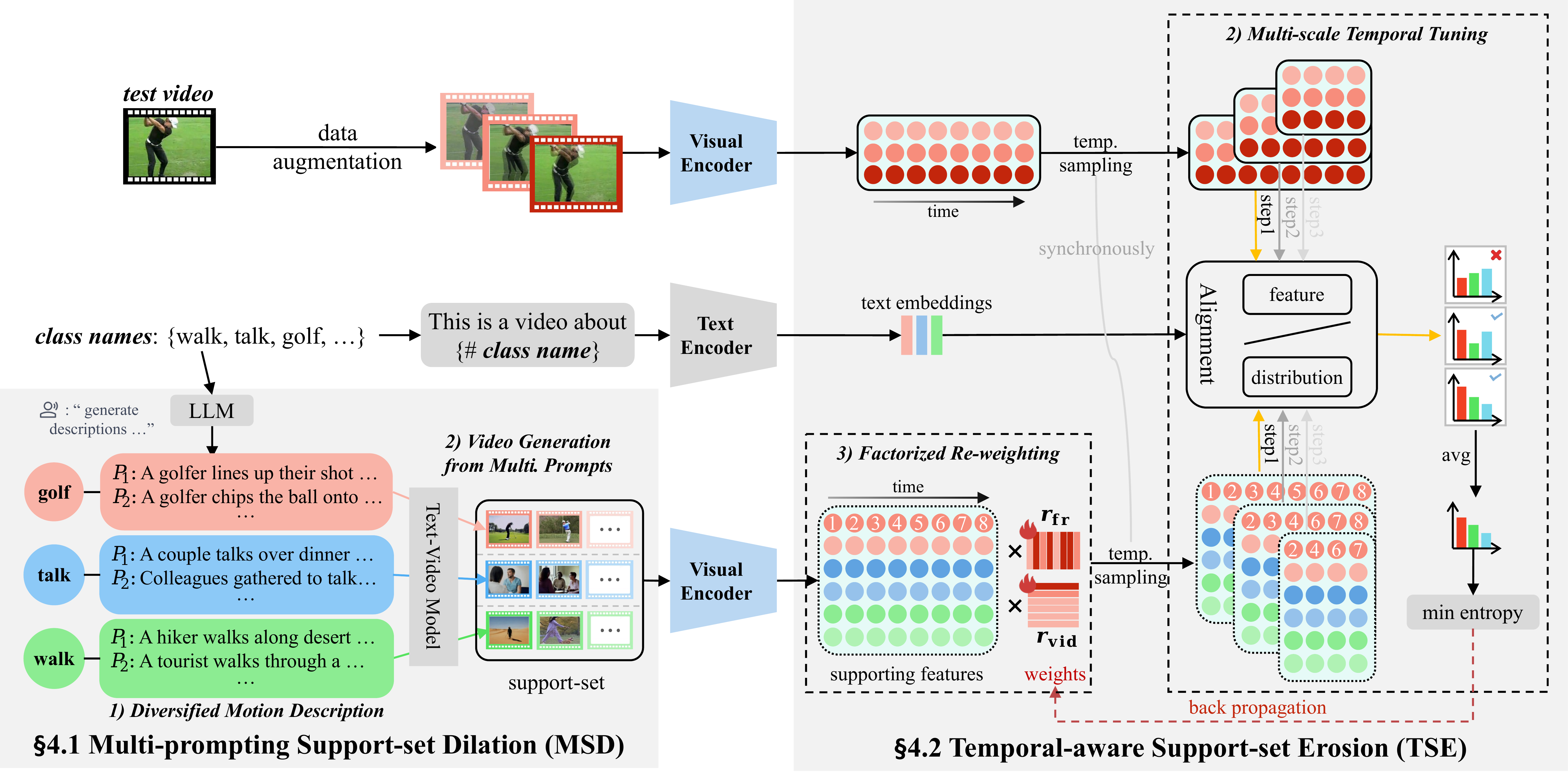}
    \caption{Overview of the proposed framework \testv~which \textit{first dilates and then erodes} the support set for zero-shot video classification.
    i) {\textbf{Multi-prompting Support-set Dilation~(MSD)}}: It builds diversified motion description for each class name via the LLM and then generates video samples with these elaborate descriptions via the text-to-video generation model for constructing a diverse support set. ii) {\textbf{Temporal-aware Support-set Erosion~(TSE)}}: Based on the visual feature of the given test video~{$\bm f$} and support set {$\bm F$}, it applies factorized weights $\bm r_{*}$ to mine critical supporting cues from the support set and tunes the weights with prediction consistency at multiple temporal scales.}\label{fig::overview}
\end{figure*}

\section{Methodology}
{In this section, we present a novel framework, \ie, TEst-time Support-set Tuning for Video classification~(\testv). It constructs a semantically diverse support-set with multiple category descriptions and tunes learnable weights to adjust the support-set with the test samples. The overview is shown in Figure~\ref{fig::overview} and details are presented below.}

\subsection{Multi-prompting Support-set  Dilation~(MSD)}\label{sec::MSD}
{As shown in Figure~\ref{fig::sus}, the performance of the support-set based on SuS-X is extremely saturated when the number of videos per category increases, suggesting that the generated video samples' representations are highly overlapping/similar in each class. This inspires this work to improve the diversity of generated samples to refine the semantic boundary of each class within the support-set. To this end, we propose the Multi-prompting Support-set Dilation~(MSD) module to generate diverse video samples via a video-text generation model with different prompts.}

\noindent\textbf{Diversified Motion Description}.
Given a set of action classes $\bm{Y}=\{\bm{y}_i\}_{i=1}^C$, we employ a large language model (LLM) (\emph{e.g.}, ChatGPT~\cite{ChatGPT}) to generate ${M}$ different prompts for each class as
\begin{equation}\label{eq::LLM}
\bm{d}_i = \texttt{LLM}(\texttt{query\_template}(\bm{y}_i,M)). 
\end{equation} 
Here, $\texttt{query\_template}(\cdot)$ is a crafted text template used to wrap each action class into a text query as the input of LLM, \textit{as shown in Figure $I$ in the Appendix}. After traversing the entire action class set $Y$, we can achieve a diversified motion description set $\bm{D}=\{\bm d_{i}\}_{i=1}^C$ in which each class corresponds to $M$ different explanations and $C$ is the number of action classes.

{
\noindent\textbf{Video Generation from Diversified Prompts}.
For the $i$-th action class, we generate $K=m \times n$ video samples via the text-to-video model $\texttt{T2V}(\cdot, \cdot)$ according to text prompts stored in $\bm d_{i}$ as follows, 
\begin{equation}
\bm{s}_i = \texttt{T2V}(\texttt{Sample}(\bm{d}_i, m), n).\label{eq::MSD}
\end{equation}
Here, we $\texttt{Sample}$ $m$ prompts for each class from $\bm{d}_i$ and repeatedly generate $n$ videos corresponding to each prompt. \textit{$K$ is set to be always factorable for ease of implementation, {two hyper-parameters (\emph{\textit{i.e.}}, $m$ and $n$) are ablated in the experiments (as shown in Figure~\ref{fig::sus})}.} After traversing the entire description set $\bm{D}$, we can construct the video support-set $\bm{S}=\{s_{i}\}_{i=1}^{C}$ in which each class has $K$ support videos.
}

{
Different from previous methods~(SuS-X~\cite{udandarao2022sus-x}, CaFo~\cite{zhang2023prompt}, and DMN~\cite{zhang2024dual}) which explain class names repeatedly, MSD constructs diverse descriptions for each class name with different context/setting/perspectives. More details can be found in Sec-B of the Appendix.
}

\subsection{Temporal-aware Support-set Erosion~(TSE)}
\label{sec::TSE}
{
The support set generated via MSD has a greater diversity in each category~(as shown in Figure~\ref{fig::vis-MSD}). MSD refines the boundary of each class but still 
contains many invalid samples~(\ie, outliers or duplications), which motivates us to further adjust the support-set, \ie, ``\textbf{eroding}" some invalid information from the set. A straightforward solution is to apply learnable weights on video samples and tune them with a self-supervision loss, inspired by TPT~\cite{shu2022tpt}. {However, video-level tuning can miss the potential fine-grained supporting cues hidden in video data. Hence, it is necessary to tune the weights on frame-level or patch-level visual features, but patch-level introduces so many parameters that test time tuning techniques cannot accomplish optimization.}
To this end, we proposed a novel Temporal-aware Support-set Erosion module to reweigh the importance of each frame of supporting videos via factorized video-frame weights, which is tuned via different temporal-scale features in multiple steps during test time. 
\begin{table*}
	\centering
	\begin{tabular}{lccccccc}
		\toprule[1pt]
		
		\textbf{Method}        & \textbf{Encoder}   & \textbf{HMDB-51}    & \textbf{UCF-101}    & \textbf{Kinetics-600}  &  \textbf{ActivityNet}   \\
		\midrule
		\rowcolor{gray!20}\multicolumn{6}{l}{\textit{Uni-modal zero-shot video recognition models}}\\
		ER-ZSAR~\cite{chen2021elaborative}                & TSM          &$35.3\pm4.6$  &$51.8\pm2.9$    &$42.1\pm1.4$ & $-$    \\
		JigsawNet~\cite{qian2022rethinking}           & R(2+1)D      &$38.7\pm3.7$  &$56.0\pm3.1$    &$-$          & $-$    \\
		\midrule
		\rowcolor{gray!20}\multicolumn{6}{l}{\textit{Adapting pre-trained CLIP}} \\ 
		Vanilla CLIP~\cite{radford2021learning}    & ViT-B/16     &$40.8\pm0.3$ &$63.2\pm0.2$     &$59.8\pm0.3$ & $-$    \\
		Vanilla CLIP$^{*}$(reproduce)           & ViT-B/16     &$39.2\pm0.2$ &$61.7\pm0.5$     &$58.4\pm0.8$      & $68.8\pm0.6$    \\
		\textbf{Vanilla CLIP +\testv~(Ours)}       & ViT-B/16     &$\mathbf{44.3\pm0.6}$ &$\mathbf{65.5\pm0.3}$     &$\mathbf{60.1\pm0.5}$      & $\mathbf{70.1\pm0.4}$\\
		\cmidrule(l{1em}r{1em}){1-6}
		ActionCLIP~\cite{wang2021actionclip}       & ViT-B/16     &$40.8\pm5.4$ &$58.3\pm3.4$     &$66.7\pm1.1$ & $-$    \\
		A5~\cite{ju2022prompting}                  & ViT-B/16     &$44.3\pm2.2$ &$69.3\pm4.2$     &$55.8\pm0.7$ & $-$    \\
		XCLIP~\cite{XCLIP}                         & ViT-B/16     &$44.6\pm5.2$ &$72.0\pm2.3$     &$65.2\pm0.4$ & $-$    \\
		Vita-CLIP~\cite{wasim2023vitaclip}         & ViT-B/16     &$48.6\pm0.6$ &$75.0\pm0.6$     &$67.4\pm0.5$ & $-$    \\
		VicTR~\cite{Kahatapitiya_2024_CVPR}        & ViT-B/16     &$51.0\pm1.3$ &$72.4\pm0.3$     &$-$          & $-$    \\
		\midrule
		\rowcolor{gray!20}\multicolumn{6}{l}{\textit{Tuning pre-trained CLIP}} \\
		BIKE~\cite{wu2023bidirectional}            & ViT-B/16     &$49.1\pm0.5$ &$77.4\pm1.0$     &$66.1\pm0.6$ & $75.2\pm1.1$    \\
		{BIKE + DTS-TPT~\cite{yan_2024_DTSTPT}}              & ViT-B/16     &${51.6 \pm 0.5}$  &${78.0 \pm 0.6}$     &${67.0 \pm 0.5}$ &${75.8 \pm0.6}$    \\
		\textbf{BIKE + \testv~(Ours)}              & ViT-B/16     &$\mathbf{52.9\pm0.6}$  &$\mathbf{78.6\pm0.5}$     &$\mathbf{68.4\pm0.4}$ &$\mathbf{76.5\pm0.5}$    \\
		\cmidrule(l{1em}r{1em}){1-6}
		ViFi-CLIP~\cite{hanoonavificlip}           & ViT-B/16     &$51.3\pm0.7$ &$76.8\pm0.8$     &$71.2\pm1.0$ & $76.9\pm0.8$    \\
		\textbf{ViFi-CLIP +\testv~(Ours)}          & ViT-B/16     &$\mathbf{53.0\pm0.9}$   &$\mathbf{78.3\pm0.3}$     &$\mathbf{73.8\pm0.6}$ & $\mathbf{78.4\pm0.5}$    \\
		\bottomrule[1pt]

	\end{tabular}
	\caption{Comparisons with state-of-the-art methods for zero-shot activity recognition.
	} 
	\label{tab:sota}
\end{table*}

\noindent\textbf{Factorized Re-weighting.}
Given the support set $\bm{S}$ generated from MSD, we utilize the pre-trained visual encoder~($\mathcal{F}_\mathrm{vis}$) from popular VLMs (\emph{e.g.}, CLIP) to extract visual features and concatenate them together as follows,
\begin{equation}
\begin{gathered}
\bm{F}_i =\mathcal{F}_\mathrm{vis}(\bm{s_{i}}), i \in [1, CK], \bm{F}_i\in\mathbb{R}^{T \times d},\\
\bm{F} = \texttt{Concat}([\bm{F}_1,\cdots,\bm{F}_{CK}]), \bm{F} \in \mathbb{R}^{CK \times T \times d}.
\end{gathered}
\end{equation}
Here, $C$, $K$, $T$, and $d$ represent the number of classes, supporting samples of each class, frames, and channels, respectively.
After that, we reweight the support-set feature $\bm{F}$ in video and frame level via factorized learnable weights as follows,
\begin{equation}
\bm{F}' = (\bm{F} \odot \bm{r}_\mathrm{vid})\odot \bm{r}_\mathrm{fr}, \bm{F}' \in \mathbb{R}^{CK \times T \times d}.\label{eq::F_re}
\end{equation}
Here, ``$\odot$" denotes element-wise matrix multiplication~(\ie, Hadamard product). $\bm{r}_\mathrm{vid} \in \mathbb{R}^{CK \times 1 \times 1}$ and $\bm{r}_\mathrm{fr} \in \mathbb{R}^{1 \times T \times 1}$ are frame-level and video-level weights, respectively. 

\noindent\textbf{Multi-scale Temporal Tuning.} 
Inspired by Test-time Prompt Tuning~(TPT~\cite{shu2022tpt}), we tune learnable weights~($\bm r$) via minimizing the prediction consistency between re-weighted support-set features and the augmented test video features as follows, 
\begin{equation}
\begin{gathered}
\min\mathcal{L}(\texttt{Pred}(\bm{F}', \{\bm{f}^\mathrm{S}_{i}\}^{n}_{i=1})).
\label{eq::single_tuning}
\end{gathered}
\end{equation}
Here $\bm{f}^{S}_{*}=\mathcal{F}_\mathrm{vis}(\texttt{Aug}^\mathrm{S}_{n}(\bm{V}_\mathrm{test}))$ denotes the augmented test video features and $\texttt{Aug}^{\mathrm{S}}_{n}(\cdot)$ performs spatial augmentation $n$ times. $\texttt{Pred}(\cdot)$ computes multi-view predictions from the support-set features and augmented test video features according to Eq.~(\ref{eq::TIP-X}). $\mathcal{L}$ calculates the averaged entropy from high-confident predictions~\cite{shu2022tpt}.

However, the temporal evolution rate of different human activities is variable, and building multi-scale representations from the temporal domain has proven beneficial in {video representation~\cite{feichtenhofer2019slowfast,yan_2024_DTSTPT}}. Therefore, Eq.~(\ref{eq::single_tuning}) can be rewritten as follows,
\begin{equation}
\min\mathcal{L}(\texttt{Pred}(\texttt{Aug}^\mathrm{T}_{m}(\bm{F}', \{\bm{f}^\mathrm{S}_{i}\}^{n}_{i=1}))).
\end{equation}Here, $\texttt{Aug}^\mathrm{T}_{m}(\cdot)$ augments two feature sets synchronously via $m$ time sampling at different temporal scales. Different sampling strategies are compared in Table~\ref{tab::MSES} and discussed in Sec.~\ref{sec::ab_temp_sampling}. \textit{Notably, this loss function is optimized in $m$ steps with different temporal-scale features~({as illustrated in Figure~$II$ in the Appendix)}.}
}
\section{Experiments}
\label{experiment}
We evaluate the effectiveness of the proposed method on four popular video benchmarks, \ie, HMDB-51~\cite{kuehne2011hmdb}, UCF-101~\cite{soomro2012ucf101dataset101human}, Kinetics-600~\cite{carreira2018shortnotekinetics600}, ActivityNet~\cite{caba2015activitynet}. Detailed configuration of the platform environment, experimental data, models, and optimization can be found in the Appendix.

\subsection{Comparison with State-of-the-Arts}
{

We evaluate the effectiveness of \testv with existing state-of-the-art methods on four popular action recognition benchmarks in Table~\ref{tab:sota}. Existing methods can be categorized into three types: 
i) Uni-modal zero-shot video recognition models: they are trained on video data with elaborated representation engineering;
ii) Adapting pre-trained CLIP: they adapt CLIP to video data via additional temporal learners or VL prompting techniques;
iii) Tuning pre-trained CLIP: they fully fine-tune the CLIP model via video data.

We apply off-the-shelf pre-trained visual/text encoders from VLMs, \eg, BIKE~\cite{wu2023bidirectional} and ViFi-CLIP, to extract features and tune the proposed \testv~during test time without any training data. Our method outperforms the conventional uni-modal zero-shot video recognition models by a large margin~($15\%\sim 20\%$). Moreover, compared with the methods adapted from pre-trained CLIP with video data, our method achieve consistent improvements across all benchmarks without training. Furthermore, our method can significantly improve some recent methods fully-tuned from CLIP, thanks to the reduction of modality gaps through the tuned support set. In general, the proposed \testv~achieves state-of-the-art performance on four popular benchmarks in a training-free manner.

}

\subsection{Ablation Study}
\paragraph{Choice of LLM and T2V for MSD.}
{
As described in Sec.~\ref{sec::MSD}, MSD builds multiple descriptions~(\ie, prompts) for each class via the Large Language Model~(LLM) and then generates video samples via the Text-to-Video~(T2V) model according to multiple prompts. There are many options for LLM and T2V, thus we compare them in Table~\ref{tab::LLM} and~\ref{tab::GEN}, and make the following analysis.
\textbf{i) LLMs}: We compare different Large Language Models (LLMs) for motion description generation defined in Eq.~(\ref{eq::LLM}) in Table~\ref{tab::LLM}. In general, our approach works well on all LLMs, and new versions of the same model perform better. ChatGPT achieves the best performance and is selected as the default LLM.
\textbf{ii) T2Vs}: The proposed method is compatible with various Text-to-Video~(T2V) generation models as shown in Table~\ref{tab::GEN}. {Benefited from the extra training videos data}, LaVie is slightly better than others and is adopted by default.
}

\begin{table}[!h]
    \centering
    \resizebox{\linewidth}{!}{
    \begin{tabular}{lcccc}
        \toprule[1pt]
        \textbf{LLM}                     & \textbf{Version}       & \textbf{HMDB-51}   & \textbf{UCF-101} \\ 
        \hline
        \multirow{2}*
        {Gemini~\cite{geminiteam2024geminifamilyhighlycapable}}   &1.0 pro               & $49.3$            & $74.5$   \\ 
                                         &1.5 pro               & $50.8$            & $76.5$   \\ 
        \midrule
        \multirow{2}*
        {Llama-3~\cite{llama3modelcard}} &8B           & $50.9$            & $76.1$     \\
                                         &70B          & $51.2$            & $77.5$     \\
        \midrule
        \multirow{3}*
        {Claude-3~\cite{Claude3}}        &haiku                 & $50.0$            & $75.5$   \\ 
                                         &sonnet                & $51.0$            & $77.1$   \\ 
                                         &opus                  & $51.1$            & $77.4$   \\ 
        \midrule
        \multirow{2}*
        {\textbf{ChatGPT}~\cite{ChatGPT}}         &\textbf{3.5-turbo}         & $\mathbf{50.1}$            & $\mathbf{77.4}$   \\
                                         &\textbf{4-turbo}           & $\mathbf{51.3}$   & $\mathbf{77.8}$   \\
        \bottomrule[1pt]
    \end{tabular}}
    \caption{Effect of different LLMs used in MSD.}
    \label{tab::LLM}
\end{table}

\begin{table}[!h]
    \centering
    \resizebox{\linewidth}{!}{
    \begin{tabular}{llc}
        \toprule[1pt]
        \textbf{\texttt{T2V} Model}   & \textbf{PT data}   & \textbf{HMDB-51}   /   \textbf{UCF-101}  \\ 
        \hline
        Show-1 & WV10M  & $50.2$ / $77.1$   \\
        HiGen                 & WV10M + Inter. data20M               & $50.3$ / $77.3$    \\
        TF-T2V                & WV10M + LAION5B + Inter. data10M     & $49.8$ / $77.3$      \\
        ModelScopeT2V         & WV10M + LAION5B                         & $51.0$ / $77.6$  \\
        \textbf{LaVie}        & WV10M + LAION5B + Vimeo25M              & $\mathbf{51.3}$ / $\mathbf{77.8}$    \\ 
        \bottomrule[1pt]
    \end{tabular}}
    \caption{Effect of different \texttt{T2V} models \tablefootnote{including Show-1~\protect\cite{zhang2023show1marryingpixellatent}, HiGen~\protect\cite{higen}, TF-T2V~\protect\cite{TFT2V}, ModelScopeT2V~\protect\cite{wang2023modelscopetexttovideotechnicalreport}, and LaVie~\protect\cite{wang2023laviehighqualityvideogeneration}} used in MSD.}
    \label{tab::GEN}
\end{table}

\begin{figure}[!t]
  \centering
  \includegraphics[width=1\linewidth]{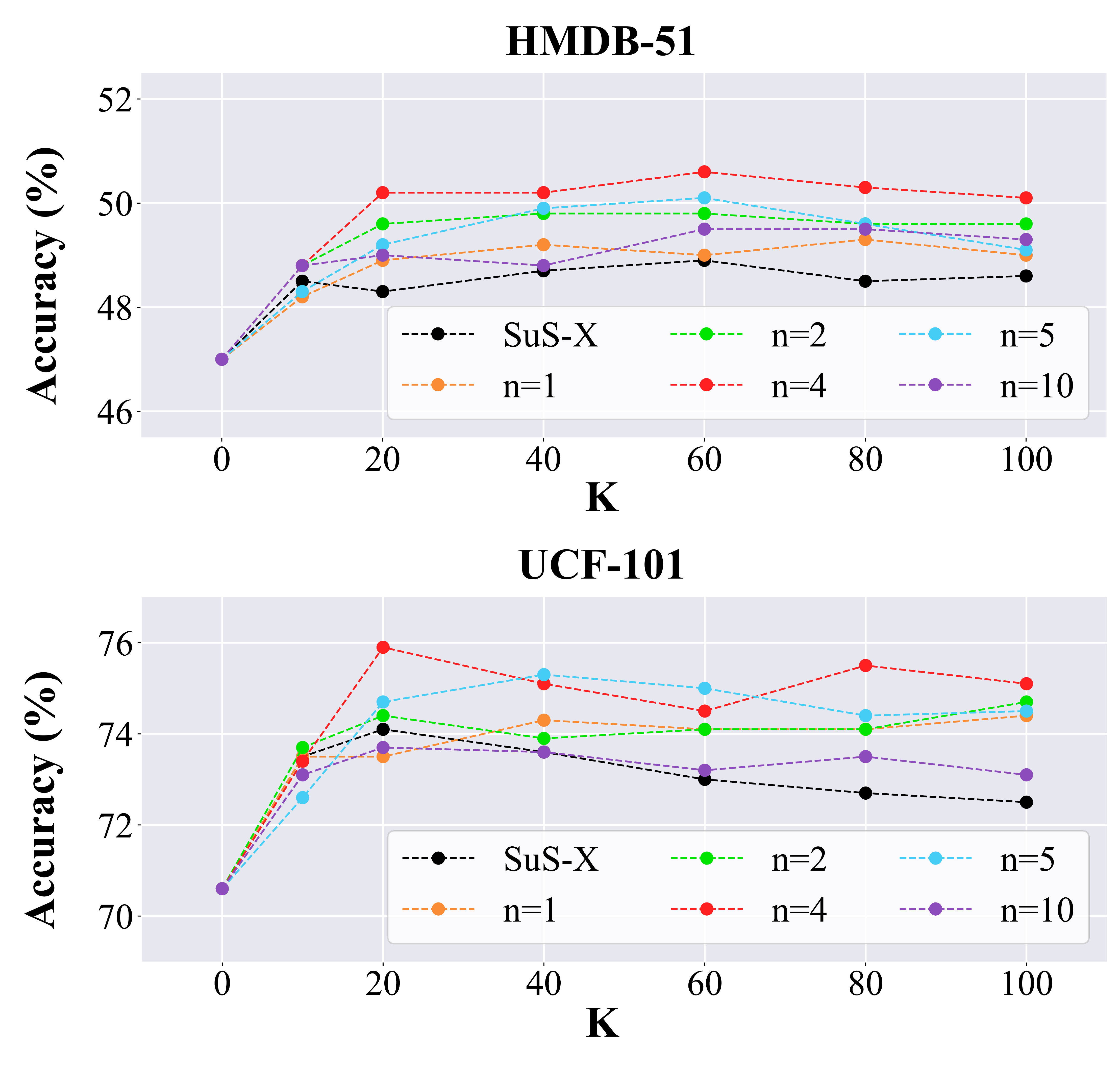}
  \caption{Effect of support hyper-parameters $K$ and $n$~($K = m \times n$ defined in the equation~\ref{eq::MSD}) with a single prompt~(SuS-X) and multiple prompts~(our MSD) on HMDB-51 and UCF-101. Top-1 zero-shot recognition accuracy is reported.}
  \label{fig::sus}
\end{figure}

\paragraph{Importance of Diversity of the Support-set}
{
    To figure out the key factor of the quality of the support-set, we ablate the support-set hyper-parameters of the proposed MSD, \ie, the number of supporting videos for each class $K$ and the number of videos repeatedly generated for each prompt $n$ defined in Equation~(\ref{eq::MSD}), as shown in Figure~\ref{fig::sus}.    
    The performance of all models~(baseline SuS-X and the proposed MSD with different $n$) gradually plateau as the support-set size~($K$) increases across two benchmarks. 
    Besides, the proposed multi-prompting method~(MSD) outperforms the single-prompting baseline~(SuS-X) when $K \neq 0$, demonstrating the importance of sample diversity in the support-set. 
    Meanwhile, we ablate the number of repeatedly generated videos $n$ when the total number of videos $K$ is fixed.~($K= m \times n$). We found that $n=4$ brings stabilized gains compared to the baseline and \textbf{set $n=4$ for all benchmarks}. The number of prompts $m$ used for each benchmark is listed in the Appendix due to limited space.
}
\begin{figure*}[ht]
  \centering
  \includegraphics[width=1\linewidth]{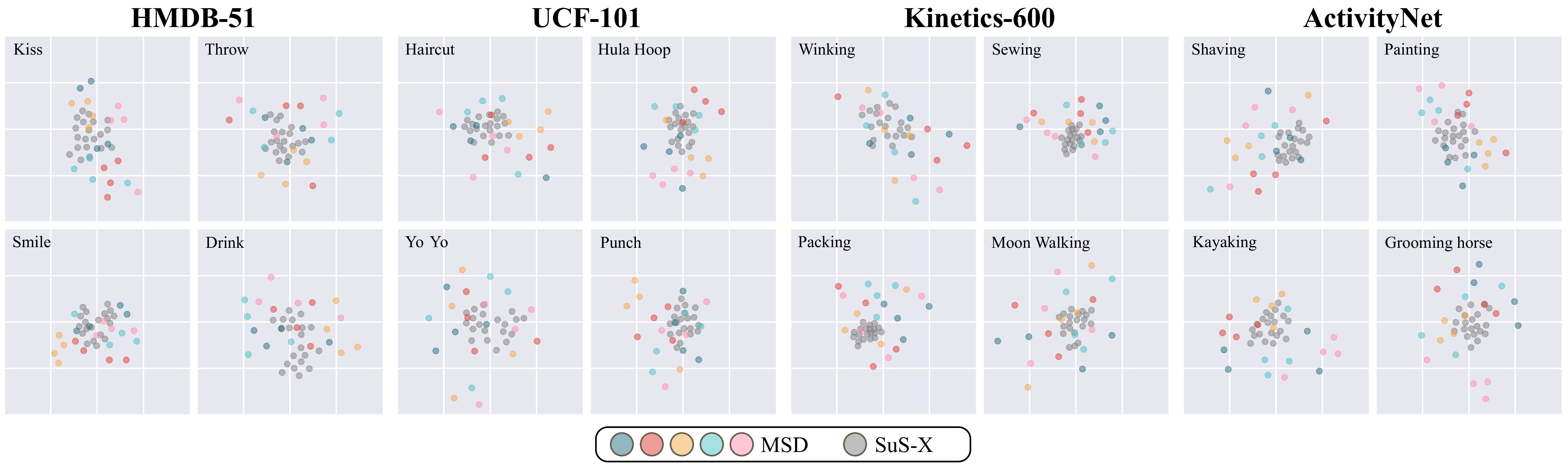}  
  \caption{Feature distribution of supporting samples generated with multiple prompts~(MSD) and single prompt~(SuS-X) on different benchmarks. Multi-prompting and single-prompting samples are shown in color and grey, respectively.}\label{fig::vis-MSD}
\end{figure*}
\begin{table}[!h]
    \centering
    \resizebox{\linewidth}{!}{
    \begin{tabular}{cccccc}
        \toprule[1pt]
        \multirow{2}*{\textbf{Dataset}} & \multirow{2}*{\textbf{Strategy}}   & \multicolumn{1}{c}{\textbf{single}} & \multicolumn{3}{c}{\textbf{multiple}}  \\ 
        \cmidrule(lr){3-3}\cmidrule(lr){4-6}
        & &\textbf{$8$} &\textbf{$6,8$}  & \textbf{$4,6,8$} & \textbf{$2,4,6,8$} \\\midrule
        \multirow{2}*
        {HMDB-51}          
                                & random           & $52.0$      & $52.0$        & $52.1$             & $51.8$      \\
                                & top              & $52.0$      & $52.6$        & $\mathbf{52.9}$    & $52.5$\\
        \midrule
        \multirow{2}*
        {UCF-101}           
                                & random           & $77.8$      & $77.7$        & $77.8$             & $78.0$\\
                                & top              & $77.8$      & $78.4$        & $\mathbf{78.6}$    & $78.4$  \\
        \bottomrule[1pt]
    \end{tabular}}
    \caption{Effect of different sampling methods for multi-scale tuning in TSE. Each video contains a total of $8$ frames.}
    \label{tab::MSES}
\end{table}

\paragraph{Multi-scale Temporal Tuning.}\label{sec::ab_temp_sampling}
{
    To find the suitable temporal scale used in TSE, we try different sampling strategies and report the performance in Table~\ref{tab::MSES}. ``random" sampling frames from the feature sequence randomly, while ``top" selects the top $k$ features with higher frame-level weights~($r_\mathrm{f}$). We observed that multi-scale tuning with the ``random" strategy does not work compared with the single-scale tuning. Besides, multi-scale tuning the ``top" strategy with ``4,6,8" achieves state-of-the-art results. 
}

\begin{table}[!h]
    \centering
    \begin{tabular}{cccc}
        \toprule[1pt]
        \multicolumn{2}{c}{\textbf{Module}} & \multirow{2}{*}{\textbf{HMDB-51}} & \multirow{2}{*}{\textbf{UCF-101}}\\ 
        \cmidrule(r){1-2}        \textbf{MSD}     & \textbf{TSE}   \\ 
        \midrule
        \xmark          &\xmark           & $50.1$                      & $76.8$   \\
        \cmark          &\xmark           & $51.3$                      & $77.8$   \\
        \xmark          &\cmark           & $51.9$                      & $78.1$   \\
        \cmark          &\cmark           & $\mathbf{52.9}$             & $\mathbf{78.6}$   \\
        \bottomrule
    \end{tabular}
    \caption{Effect of different components of our approach.}
    \label{tab::Comp}
\end{table}
\begin{table}[!ht]
    \centering
    \begin{tabular}{ >{\centering\arraybackslash}p{1cm} >{\centering\arraybackslash}p{1cm} cc}
        \toprule[1pt]
        \multicolumn{2}{c}{\textbf{Learnable weights}} & \multirow{2}{*}{\textbf{HMDB-51}} & \multirow{2}{*}{\textbf{UCF-101}}  \\  
        \cmidrule(r){1-2}    $\bm{r}_\mathrm{vid}$     & $\bm{r}_\mathrm{fr}$    \\ 
        \midrule
          
      \xmark    &\xmark       & $51.3$              & $77.8$      \\  
                \cmark    &\xmark       & $52.1$              & $78.3$     \\    
                \xmark    &\cmark       & $52.4$              & $78.0$     \\  
            \cmark    &\cmark       & $\mathbf{52.9}$     & $\mathbf{78.6}$  \\  

        \bottomrule
    \end{tabular}
    \caption{Effect of different components in weights module.} 
    \label{tab::WM}
\end{table}

\paragraph{Component Analysis.}
{
    We explore the effect of each module designed in \testv~in Table~\ref{tab::Comp}. Compared with the baseline, MSD brings approximately $1\%$ improvements on two benchmarks, indicating that the diversity of supporting samples in each class is beneficial in enhancing the quality of the support set. Meanwhile, TSE demonstrates the ability to select critical visual cues from highly redundant video data, which improves the baseline by approximately $1\%\sim 2\%$. Furthermore, the combination of MSD and TSE yields even more significant performance gains, confirming the complementarity of the two modules.

    As defined in Eq.~(\ref{eq::F_re}), learnable weights $\bm{r}_\mathrm{vid}$ and $\bm{r}_\mathrm{fr}$ are designed to adjust the contribution of video-level and frame-level features, respectively. As reported in Table~\ref{tab::WM}, each of them brings performance gains and these two weights are complementary.
}

\begin{table}[ht]
    \centering
    \resizebox{\linewidth}{!}{
    \begin{tabular}{llcc}
    \toprule[1pt]
    \textbf{Backbone}              &\textbf{Method}              & \textbf{HMDB-51} & \textbf{UCF-101} \\ 
    \hline
    \multirow{6}*
    {ViT-B/16}  &Vanilla CLIP                    & $39.2$             & $61.7$   \\
                &+ SuS-X                         & $41.7$             & $64.5$   \\
                &\textbf{+ \testv~(Ours)}               & $\mathbf{44.3}$         & $\mathbf{65.5}$     \\ 
    \cmidrule(rll){2-4}
                &BIKE                            & $47.0$           & $70.6$  \\
                &+ SuS-X                         & $50.1$           & $76.8$   \\
                &\textbf{+ \testv~(Ours)}                & $\mathbf{52.9}$        & $\mathbf{78.6}$   \\ 
    \midrule
    \multirow{6}*
    {ViT-L/14}  &Vanilla CLIP                    & $45.9$             & $69.8$   \\
                &+ SuS-X                         & $46.8$             & $73.6$   \\
                &\textbf{+ \testv~(Ours)}                & $\mathbf{48.2}$             & $\mathbf{75.7}$     \\ 
    \cmidrule(rll){2-4}
                &BIKE                            & $53.2$           & $80.5$  \\
                &+ SuS-X                         & $56.7$           & $85.0$   \\
                & \textbf{+ \testv~(Ours)}                & $\mathbf{59.7}$             & $\mathbf{87.5}$   \\ 
    \bottomrule[1pt]

\end{tabular}}
    \caption{Generalization to different pre-trained VLMs.
    }
    \label{tab::PM}
\end{table}

\paragraph{Generalization to Different VLMs.}
{
    We also extract visual/text features via different pre-trained VLMs, \ie, Vanilla CLIP~\cite{radford2021learning}, and BIKE~\cite{wu2023bidirectional}, 
    with various network backbones,
    and report top-1 zero-shot accuracy on two benchmarks in Table~\ref{tab::PM} compared with SuS-X~\cite{udandarao2022sus-x}. Our approach shows consistent improvement across two benchmarks with different models and backbones, demonstrating good generalization across different video representations.   
}

\paragraph{Visualization.}
{
    We randomly sample $20$ videos from the support sets generated with multiple prompts and the single prompt for four benchmarks, and all videos are embedded via the pre-trained visual encoder of BIKE~\cite{wu2023bidirectional} with $4$ frames. After that, the feature distributions of sampled videos in each benchmark are visualized via t-SNE~\cite{van2008visualizing} in Figure~\ref{fig::vis-MSD}. We observed that single-prompting samples are more likely to cluster together, whereas other multi-prompting samples are more dispersed. This phenomenon shows that the multi-prompting strategy can refine the supporting boundary to enhance the quality of the support-set.
    

}
\section{Conclusion}
\label{conclusion}
{This work presents a novel framework, namely \testv~which \texttt{dilates} and \texttt{erodes} the support set to enhance the zero-shot generalization capability of video classification methods during test time. \testv~applies multiple prompts to enrich the support set~(\texttt{Dilation}) and then mines critical supporting samples from the support set with learnable spatio-temporal weights~(\texttt{Erosion}). Our method demonstrated superior performance to existing state-of-arts on four benchmarks. While \testv~effectively adapts pre-trained models~(i.e., CLIP) to out-of-distribution domains during test time for video data, it still has two potential limitations: \textbf{i)} extra spatio-temporal tuning costs may increase with the length of the video~(i.e., long video). \textbf{ii)} relying on the quality of LLMs and text-to-video generation models.


}

\section*{Acknowledgments}
      {The work is supported by the National Natural Science Foundation of China (Grant No. 62472208), the Postdoctoral Fellowship Program of CPSF (Grant No. GZB20230302), the China Postdoctoral Science Foundation (Grant No. 2023TQ0151 and 2023M731596), the Jiangsu Funding Program for Excellent Postdoctoral Talent (Grant No. 2023ZB256). Rui Yan and Jin Wang equally contributed to this work, and Tieniu Tan is the corresponding author.}

\bibliographystyle{named}
\bibliography{ijcai25}

\newpage
\appendix
\onecolumn
 This document provides more details of our approach, which are organized as follows:
 {
\begin{itemize}
 \item \S A provides the overview of the MSD and TSE modules.
 \item \S B provides the discussion on the Speed and Memory of the proposed method.
 \item \S C provides the detailed configuration used for the experiment.
 \item \S D provides the visualization of Support-set Construction.
\end{itemize}
}

\section{Overview of MSD and TSE Module}
{
\subsection{Support-set constructed in MSD}
\noindent\textbf{Overview.}~{
To further reveal the process of the Multi-prompting Support-set Dilation (MSD), we illustrate the various steps of these modules in Figure~\ref{fig::MSD}. We use a query template to transform each class name into detailed prompts using a large language model (LLM), \eg, ChatGPT~\cite{ChatGPT} and Claude~\cite{Claude3}. Each generated prompt is then fed into the text-to-video model, \eg, ModelScopeT2V~\cite{wang2023modelscopetexttovideotechnicalreport}, and LaVie~\cite{wang2023laviehighqualityvideogeneration} for generating video samples. 
}

\begin{figure}[!ht]
  \centering
  \includegraphics[width=1\linewidth]{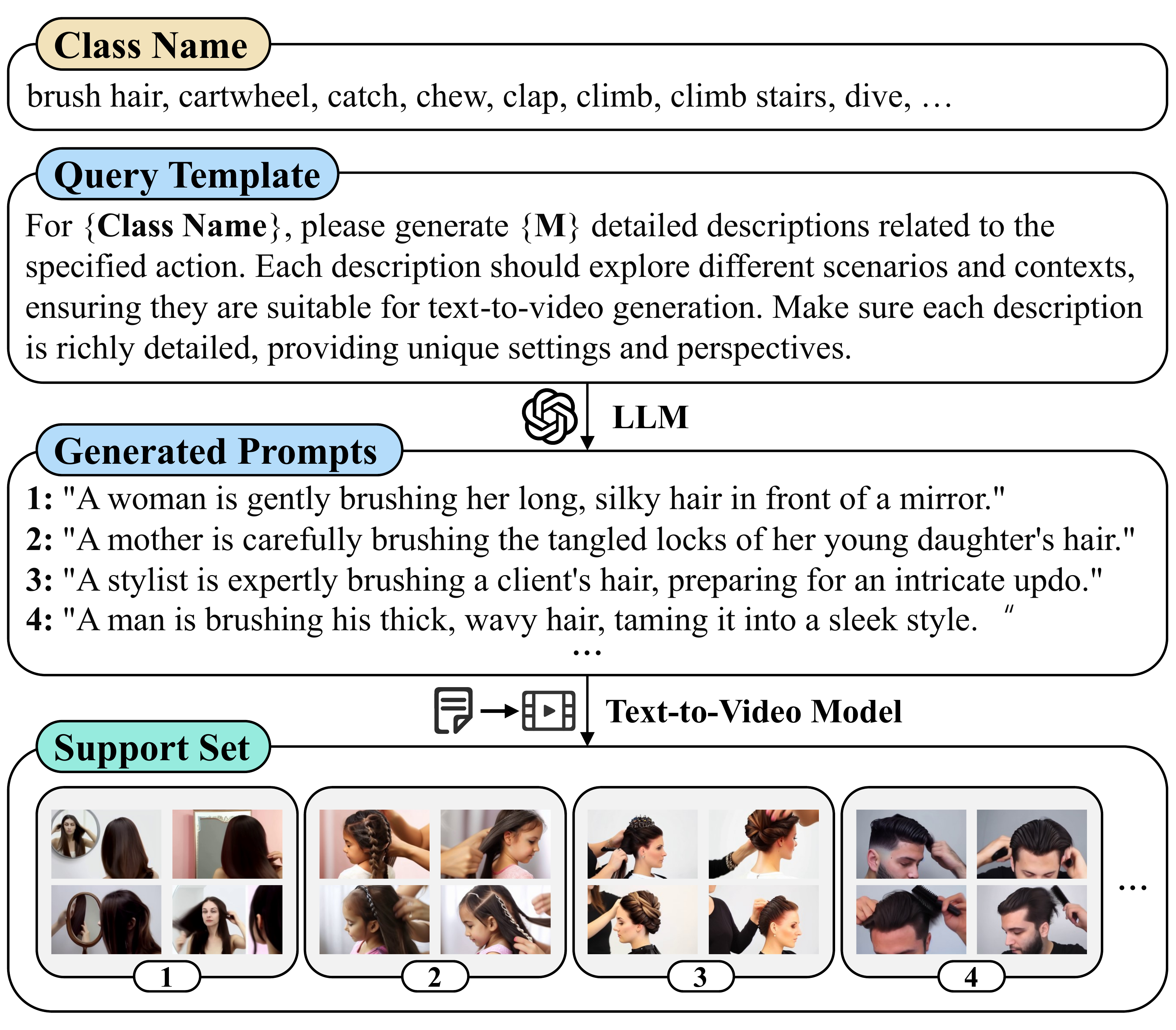}
  \caption{Multi-prompting Support-set Dilation (MSD).}\label{fig::MSD}
\end{figure}

\noindent\textbf{Discussion.}~{
SuS-X~\cite{udandarao2022sus-x}, CaFo~\cite{zhang2023prompt}, and DMN~\cite{zhang2024dual} apply different language commands to describe a class as clearly as possible via LLMs. (e.g., ``What a [CLASS] looks like?", and ``How can you identify a [CLASS]?"). They can be treated as ``\textbf{Explanators}". However, generative models often cannot depict the fine details of human motion but seem good at fabricating some inessential contextual information. Thus, MSD applies only one command and focuses on constructing diverse textual descriptions of motion with different contexts/settings/perspectives, rather than repeatedly explaining the motion, as shown in Figure~\ref{fig::MSD} of Supplementary Materials. Hence, MSD can be treated as a ``\textbf{Constructor}".
}

\begin{figure}[!htbp]
  \centering
  \includegraphics[width=1.0\linewidth]{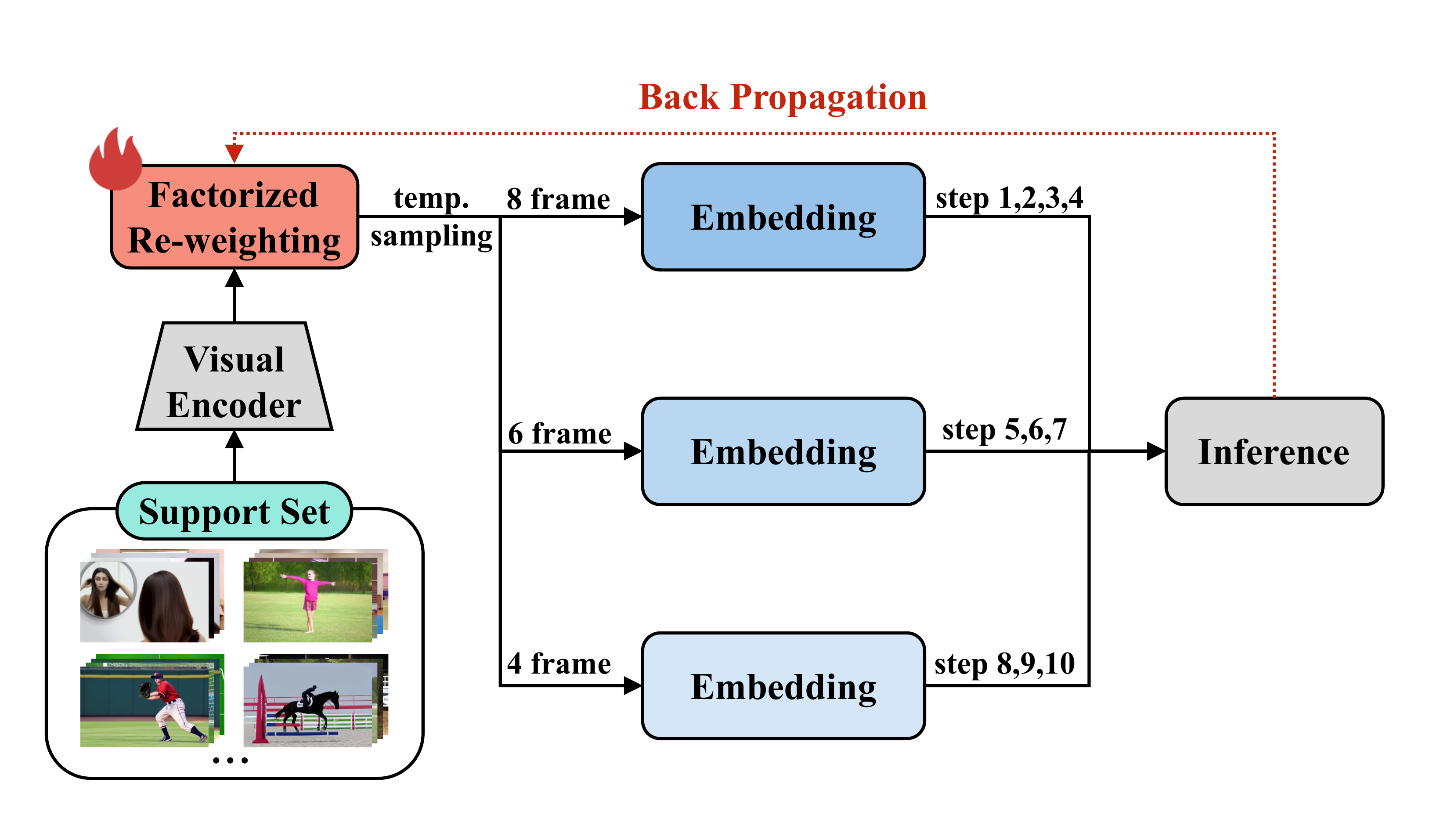}
  \caption{Temporal-aware Support-set Erosion (TSE).}\label{fig::TSE}
\end{figure}

\subsection{Multi-scale Temporal Tuning in TSE}
{
We illustrate the process of multi-scale temporal tuning in Temporal-aware Support-set Erosion (TSE) modules in Figure~\ref{fig::TSE}. For the TSE module, we tune the weights with different temporal scale visual features extracted from pre-trained Vision-Language Models~(VLMs), \eg, BIKE~\cite{wu2023bidirectional} and ViFi-CLIP~\cite{hanoonavificlip}, in multiple steps. Specifically, given the multi-scale features with ``4, 6, 8" frames, we tune the weights with $8$ frames in four steps~(``1,2,3,4"), with $6$ frames in three steps~(``5,6,7"), and with $4$ frames in three steps~(``8,9,10"). 
}
}

\begin{table*}[!t]
    \centering
    \begin{tabular}{ccccc}
        \toprule[1pt]
        \textbf{Dataset}      & \textbf{$\#$ class (C)}      & \textbf{$\#$ text prompt} & \textbf{$\#$ videos for each prompt ($n$)} & \textbf{support-set size}\\ 
        \hline
        HMDB-51     & $51$      & $15$  & $4$    & $51 \times 15 \times 4$\\
        UCF-101     & $101$     & $5$   & $4$    & $101 \times 5 \times 4$\\
        Kinetics-60 & $220$         & $4$   & $4$    & $220 \times 4 \times 4$\\
        ActivityNet & $200$       & $6$   & $4$    & $200 \times 6 \times 4$\\
        \bottomrule[1pt]
    \end{tabular}
    \caption{Model configuration of prompt and support-set for each benchmark.}
    \label{tab::model_config}
\end{table*}

\section{Discussion on Speed and Memory}{

\subsection{Multi-prompting Support-set Dilation~(MSD)}
The support-set is built offline and the time cost varies on the choice of Large language Models~(LLMs) and Text-to-Video models (T2Vs). LLMs are accessed via API, thus the computational cost is unknown and the corresponding processing time depends on the network speed. GPU memory and processing times of T2Vs for one single video are shown in Table~\ref{tab::infer_time_t2v}.
\begin{table}[!h]
    \centering
    \begin{tabular}{lcc}
        \toprule[1pt]
        \textbf{T2V}      & \textbf{GPU Mem}      & \textbf{Time} \\ 
        \hline
        Show-1~\cite{zhang2023show1marryingpixellatent}      & $16$GB     & $~60$s     \\
        HiGen~\cite{higen}       & $16$GB     & $~40$s     \\
        TF-T2V~\cite{TFT2V}      & $15$GB     & $~40$s     \\
        ModelScopeT2V~\cite{wang2023modelscopetexttovideotechnicalreport}  & $15$GB  & $~20$s     \\
        LaVie~\cite{wang2023laviehighqualityvideogeneration}       & $14$GB     & $~20$s     \\
        \bottomrule[1pt]
    \end{tabular}
    \caption{Inference costs of T2Vs.}
    \label{tab::infer_time_t2v}
\end{table}

\subsection{Temporal-aware support-set Erosion~(TSE)}
As shown in Table~\ref{tab::infer_time}, despite some parameters and memory introduced by \testv, the sparse frame sampling strategy allows \testv~to understand $3 \sim 5$-sec videos (such as those in HMDB-51 and UCF-101) in less than 1 second. Therefore, we believe that such inference time can still meet the real-time requirements of video understanding.
\begin{table}[!h]
    \centering
    \begin{tabular}{ccccc}
        \toprule[1pt]
        \textbf{Method}      & \textbf{Time}      & \textbf{Params} & \textbf{Mem}  & \textbf{ACC}\\ 
        \hline
        SuS-X     & $~0.6$s     & $-$     & $6$GB     & $50.1\%$\\
        \testv    & $~0.9$s     & $3.5$K   & $8.1$GB       & $\mathbf{52.9\%}$\\
        \bottomrule[1pt]
    \end{tabular}
    \caption{Statistics of inference cost on the single test video with ViT-B/16 based BIKE~\protect\cite{wu2023bidirectional} and overall accuracy on HMDB-51.}
    \label{tab::infer_time}
\end{table}


}

\section{Experimental Configuration}
\subsection{Environment}
{All experiments in this study were conducted in an operating environment consisting of an NVIDIA GeForce RTX 3090 GPU with 24 GB of VRAM, an 8-core CPU, and 32 GB of RAM. The software environment included Ubuntu 20.04, Python 3.8, and PyTorch 2.0.1, which enabled efficient training and inference of our models.}

\subsection{Datasets}
\begin{itemize}
    \item \textbf{HMDB-51}~\cite{kuehne2011hmdb} collects videos from multiple sources, such as movies and online videos. The dataset consists of a total of $6,766$ video clips from $51$ action categories, where each category contains at least $101$ video clips. We adopt three default \texttt{test} splits provided by~\cite{kuehne2011hmdb} and report the mean accuracy.
    \item \textbf{UCF-101}~\cite{soomro2012ucf101dataset101human} collects a total of $13,320$ action video clips in $101$ categories from YouTube. Each clip has a consistent frame rate of $25$ FPS and a resolution of $320 \times 240$ and is annotated into $5$ different types of categories: Body motion, Human-human interactions, Human-object interactions, Playing musical instruments, and Sports. We adopt three default \texttt{test} splits provided by~\cite{kuehne2011hmdb} and report the mean accuracy.
    \item \textbf{Kinetics-600}~\cite{carreira2018shortnotekinetics600} is a large-scale action recognition dataset that collects a total of $480$K video clips of $600$ action categories from YouTube. Each clip is manually annotated with a singular action class and lasts approximately $10$ seconds. Following~\cite{chen2021elaborative}, we choose the 220 new categories outside of Kinetics-400~\cite{kay2017kinetics} in Kinetics-600 and then split them into three splits for evaluation. The mean accuracy of these splits is reported.
    \item \textbf{ActivityNet}~\cite{caba2015activitynet} contains $200$ different types of activities and a total of $849$ hours of videos collected from YouTube. This dataset comprises $19,994$ untrimmed video clips and is divided into three disjoint subsets~(\ie, \texttt{train/val/test}) by a ratio of $2:1:1$. Following~\cite{yan_2024_DTSTPT}, we evaluate the proposed method on the \texttt{val} set.
\end{itemize}

\subsection{Implementation Details}
{
    We adopt LaVie~\cite{wang2023laviehighqualityvideogeneration} to generate video samples in Multi-prompting Support-set Dilation~(MSD) module. If not specified, the visual and text encoders are both inherited from CLIP with ViT-B. 
    For Temporal-aware Support-set Erosion (TSE), we sample $8$ frames from the test video and utilize AugMix~\cite{hendrycks2020augmix} to augment them into $32$ views. 
    Finally, we initialize all learnable weights~($\bm{r}_\mathrm{vid}$ and $\bm{r}_\mathrm{fr}$) to $1$ and tune them via AdamW optimizer~\cite{loshchilov2019decoupledweightdecayregularization} with a learning rate of $0.001$. We show the detailed configurations of the proposed methods, such as the number of epochs, the size of the textual prompt, and the size of the support set in Table~\ref{tab::model_config}. Notably, \testv, is tuned within $3$ epochs for each video with different temporal scales: (4,6,8).

}

\section{Visualization of Support-set Construction}
\begin{figure*}[!ht]
  \centering
  \includegraphics[width=\textwidth]{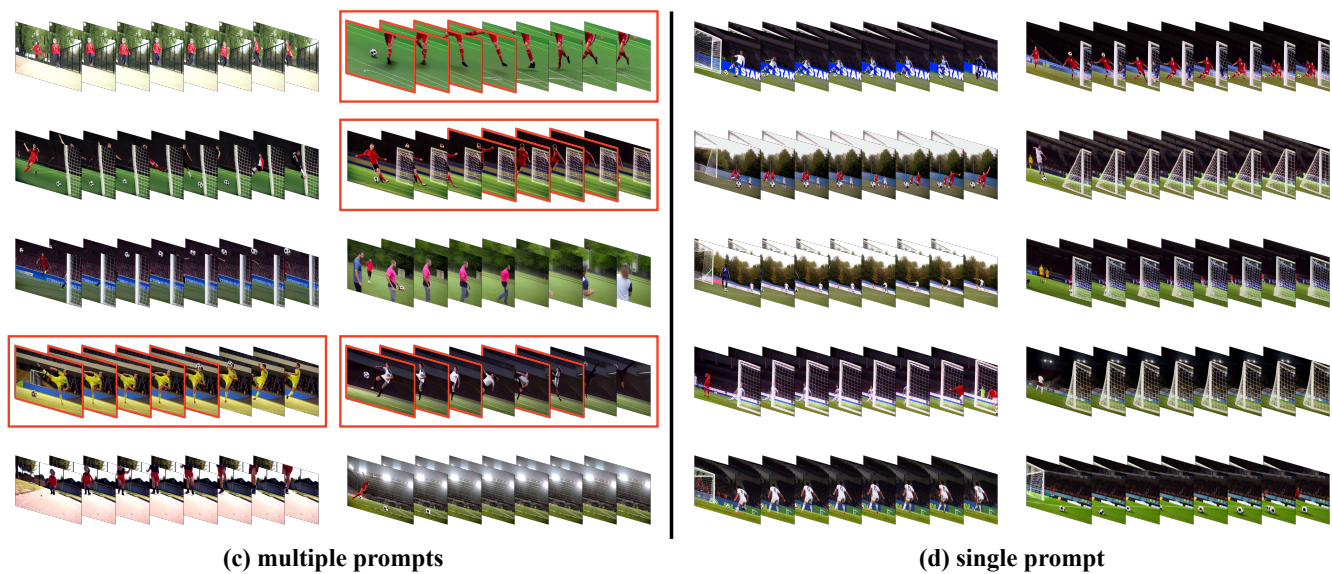}  
  \caption{Visualization of frame level samples in the support set eroded by the TSE module on HMDB-51 dataset. Multiple prompts generated videos showcase the actions of ``golf" and ``kick ball" from various backgrounds and viewpoints. For example, in ``golf", the players hit the ball from different angles, and both indoor and outdoor backgrounds are included in ``kick ball”. In addition, the samples chosen by TSE are more action-related and diverse: the players in ``Golf" wearing different colors, and the athletes in ``kick ball” shoot in different poses.}\label{fig::vis_hmdb}
\end{figure*}

\begin{figure*}[!ht]
  \centering
  \includegraphics[width=\textwidth]{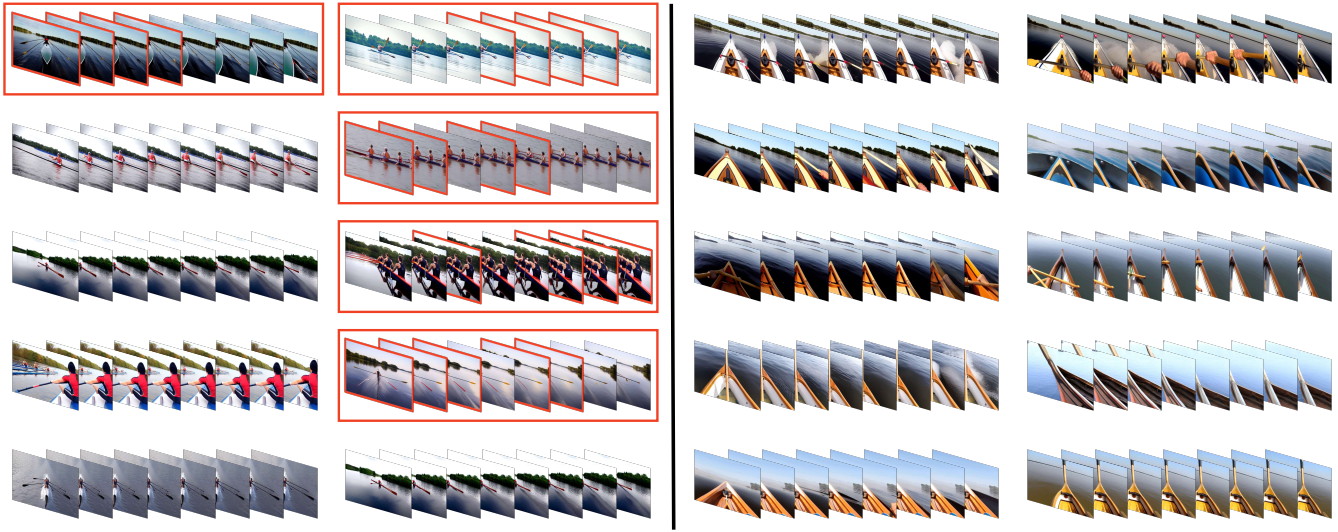}  
  \caption{Visualization of frame level samples in the support set eroded by the TSE module on UCF-101 dataset. The multiple prompts generated videos demonstrate the cutting action of various objects (potato, onion, carrot, ...) in ``Cutting In Kitchen" and illustrate the different viewpoints of the boat in ``Rowing". Furthermore, the samples captured by TSE showcase different cutting directions in ``Cutting In Kitchen" and select different views in ``Rowing" without duplication.}\label{fig::vis_ucf}
\end{figure*}

\begin{figure*}[!ht]
  \centering
  \includegraphics[width=\textwidth]{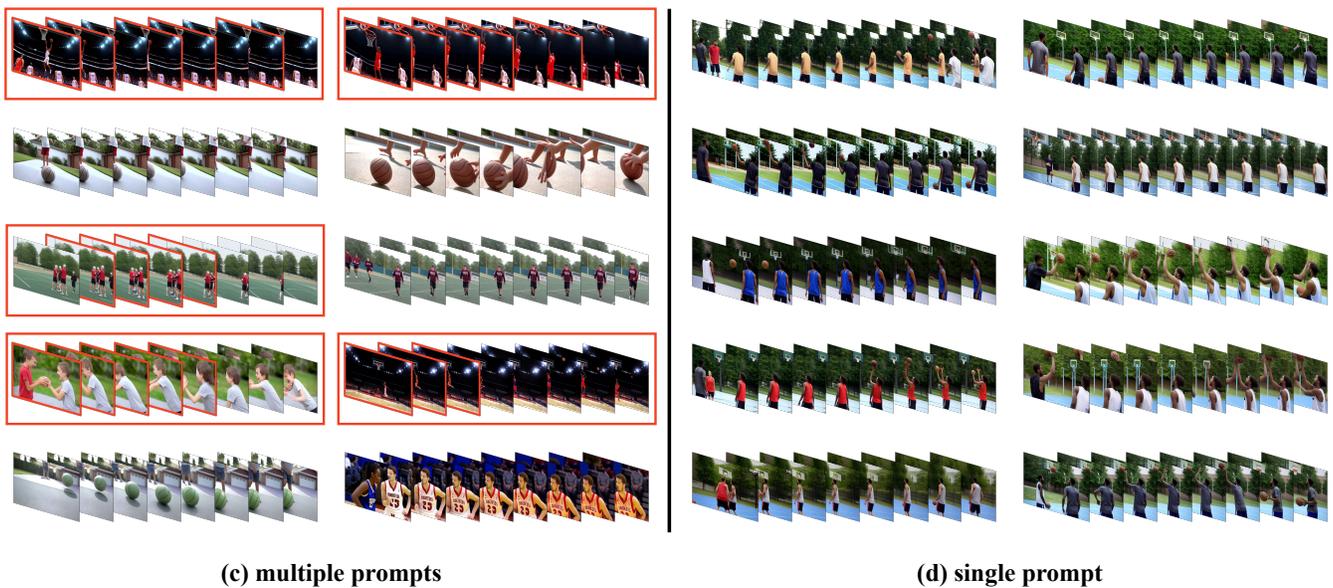}  
  \caption{Visualization of frame level samples in the support set eroded by the TSE module on Kinetics-600 dataset. The samples generated by multiple prompts, which contained different backgrounds compared to single prompt, as evidenced by the different water environments (lake, sea) in ``Catching Fish" and varied venues (indoor, outdoor) in ``Playing Basketball". Furthermore, TSE tends to select frames and samples that are motion-related: coherent catching actions in ``Catching Fish" as well as those with prominent body actions in ``Playing Basketball".}\label{fig::vis_k600}
\end{figure*}

\begin{figure*}[!ht]
  \centering
  \includegraphics[width=\textwidth]{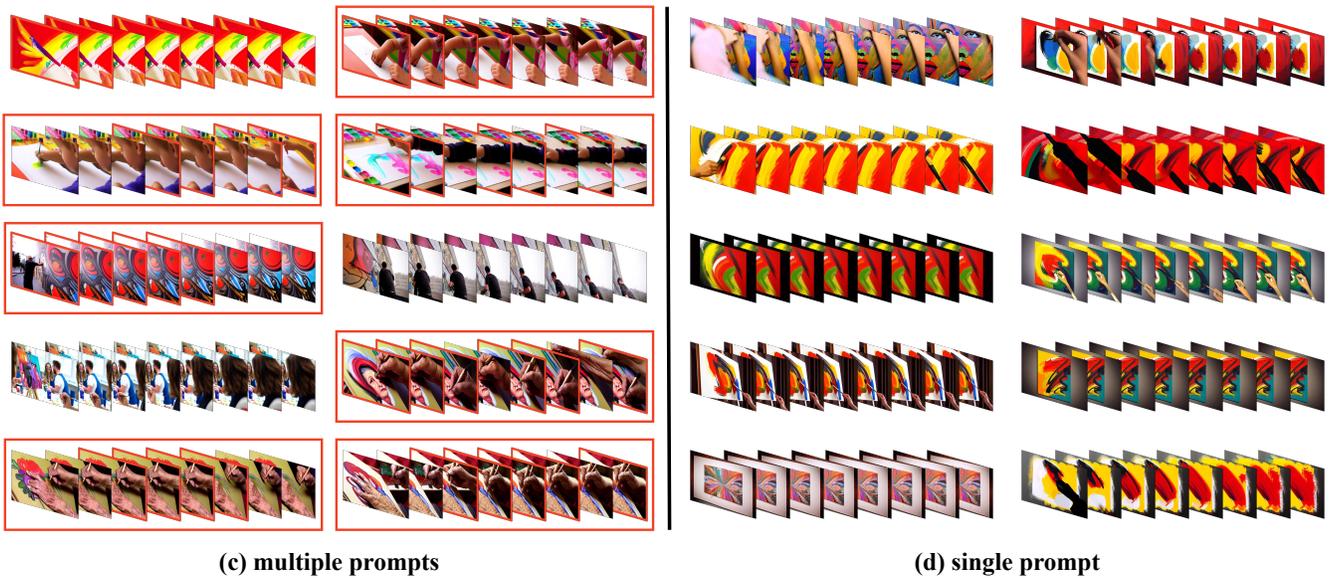}  
  \caption{Visualization of frame level samples in the support set eroded by the TSE module on ActivityNet dataset. By comparing the samples generated by multiple prompts and the single prompt, ``Applying Sunscreen" includes actions of applying sunscreen from various viewpoints (front, back) and in different backgrounds (beach, outdoor), while ``Painting" demonstrates painting on different objects (canvas, wall). Moreover, TSE is more likely to select samples with clear and coherent motions, such as the consistent application in ``Applying Sunscreen", the hand movement, and the application of paint in ``Painting".}\label{fig::vis_anet}
\end{figure*}

{
We select two action categories from each benchmark and feed them in MSD for generating $10$ supporting videos, as shown in Figure~\ref{fig::vis_hmdb}, Figure~\ref{fig::vis_ucf}, Figure~\ref{fig::vis_k600}, and Figure~\ref{fig::vis_anet}. \textit{Multi-prompting generated videos are more diversified in backgrounds, objects, clothes, and viewpoints.} In addition, we highlight the videos and frames given higher weights by TSE in red boxes~(as shown in Figure~\ref{fig::vis_hmdb}~(a) and (c), Figure~\ref{fig::vis_ucf}~(a) and (c), Figure~\ref{fig::vis_k600}~(a) and (c), and Figure~\ref{fig::vis_anet}~(a) and (c)). \textit{We can observe that support cues captured by TSE are motion-related and low redundant~(visually non-repetitive).} Detailed explanations can be found in the caption of each figure.




}


\end{document}